# Robustness Tests of NLP Machine Learning Models: Search and Semantically Replace

Rahul Singh[1], Karan Jindal[1], Yufei Yu[1], Hanyu Yang[1], Tarun Joshi, Matthew A. Campbell, Wayne B. Shoumaker

Corporate Model Risk, Wells Fargo, USA

## Abstract

This paper proposes a strategy to assess the robustness of different machine learning models that involve natural language processing (NLP). The overall approach relies upon a Search and Semantically Replace strategy that consists of two steps: (1) Search, which identifies important parts in the text; (2) Semantically Replace, which finds replacements for the important parts, and constrains the replaced tokens with semantically similar words. We introduce different types of Search and Semantically Replace methods designed specifically for particular types of machine learning models. We also investigate the effectiveness of this strategy and provide a general framework to assess a variety of machine learning models. Finally, an empirical comparison is provided of robustness performance among three different model types, each with a different text representation.

## 1 Introduction

Machine learning models used in the real world often operate under dynamically changing environments that cause degradation of model performance. Thus, during model development, one is not only interested in developing a model with the best performance (in terms of static data), but also in a model that is robust (i.e., with minimum performance degradation) under different operating conditions. Perturbations in the data are commonly encountered that can create an adversarial effect on a machine learning model and reduce its effectiveness. An adversarial attack on a machine learning model is a process for generating such adversarial perturbations. For example, in financial institutions, these operating conditions such as financial crimes or misconduct are common among models and we need advance methods to capture such events [1]. Sophisticated machine learning models, such as deep neural networks (DNNs), have been shown to exhibit a striking vulnerability to adversarial testing and lack robustness. In text classification problems, small perturbations can be designed (see [2, 3]) that can change a model's decision. These adversarial tests can be used to detect a machine learning model's robustness by measuring the drop in performance when they are applied. Adversarial attacks have also been shown to degrade the performance of other machine learning models like support vector machines (SVM) [4] and tree-based ensemble models [5]. In these examples, small perturbations in out-of-sample data cause substantial performance drops in different types. Such tests can be used to create more robust models through a process called adversarial training [3]. With NLP, the inputs are textual data and hence, ideas explored in this field relate to perturbations in the textual domain. Examples of textual perturbation include changes at the character level [6], word level [7, 8], and sentence level [9]. Adversarial conditions affect various types of NLP tasks including, for example, text classification [6, 8], machine translation [9], and question answering [10].

---

[1] All authors contributed equally.

Table 1.1: Examples of the attack strategy "Search and Semantically Replace"

| Search | Semantically Replace | | Final Attack |
|---|---|---|---|
| | Replace | Semantic Constraint | |
| The device is easy to use, but selection of a station to listen to with **good** reception is difficult.<br><br>Important token: good | The device is easy to use, but selection of a station to listen to with [**fantastic, fine, adequate**] reception is difficult.<br><br>Replacements: fantastic, fine, adequate | The device is easy to use, but selection of a station to listen to with **fantastic** reception is difficult.<br><br>**Constraints**: **fantastic** | The device is easy to use, but selection of a station to listen to with **fantastic** reception is difficult.<br><br>Attack: fantastic |

Initial efforts in adversarial testing were focused on creating changes and introducing noise in textual data. Although character level changes, adding or deleting words, etc., can create adversarial text, these actions tend to change the meaning of text and can be easily detected by humans. Synonym replacement [9] provides a better option to create text that preserves semantics; for example, with TextFooler [11]. However, word level similarity does not necessarily imply text level similarity. For example, synonym replacement can still change the semantic content of the text. Recently, more options have been explored using language models that can maintain the meaning of text. It has been shown that language models are more accurate in replacing text and creating adversarial text [12, 13, 14]. Taking this into consideration, we introduce a strategic process to target a wide range of machine learning models depending on the text representation. We present an example of the attack on a body of text in Table 1.1. The example shows the steps, (1) Search: find important tokens in the text, (2) Semantically Replace, Replace: find meaningful replacements, and Semantic Constraint: find the most meaningful component. The whole methodology is presented in more detail in the section below as Search and Semantically Replace Approach.

The main contributions from the paper are as follows:

- A set of techniques to apply different types of text representations to different types of machine learning models;
- A recipe to determine the optimal attack for a particular type of machine learning model; and
- Different types of white-box and black-box attacks designed specifically for different types of machine learning models.

We focus on text classification problems that are very common in financial applications, and on a class of adversarial attacks known as white-box attacks. White-box attacks refer to situations where information is available about the model, such as weight parameters of the model. Another, more interesting, class is known as black-box attacks, where nothing is known about the model, including the information on architecture and the internal weight parameters of the model. The only information available is the final model predictions for certain model inputs.

The paper is divided into five sections. Section 2 describes the methodology of attack strategies, followed by simulation experiments performed on a public complaint dataset. The testing data and models are described in Sections 3 and 4 respectively, followed by results and discussions in Section 5 and conclusion in Section 6.

## 2 Methodology

In regulated business environments like the financial industry, traditional machine learning models such as logistic regression, SVM, extreme gradient boosting (XGBoost), and smaller DNN architectures such as long short-term memory (LSTM), convolution neural network (CNN) are very popular. Here we focus on these machine learning models, although the methods can be translated easily to transformer-based architectures [15]. Stated broadly, a text classification problem can be organized into one of the categories as shown in Figure 2-1.

We study three types of models based on how text is represented in numeric format. The three representations are:

1. Sparse Vector representations – It is a representation in which the input is represented as a high-dimensional sparse vector. Each value in the vector represents a word; a word's representation vector has all zero values except the index of the word, which is filled with some non-zero value. The following are some of the sparse vector representations:
    a. Bag of words (BoW) – The value in the vector is filled with the frequency of the word in a single document.
    b. Term frequency-inverse document frequency (Tf-idf) – The text is represented in the form of a vector equal to the size of the vocabulary of text disregarding the grammar and semantics of text. The values in the vector are filled with tf-idf values.
    $$tfidf_{w,d} = TF_{w,d} \circ \log \frac{D}{D_w}$$
    $TF_{w,d}$ represents the frequency of word $w$ in document $d$. $D$ represents the total number of documents and $D_w$ represents total number of documents in which word $w$ occurs.
2. Word Vectors – Most of the DNN models require text to be in the form of 2D vector.
    $$v_d = \begin{bmatrix} \bar{v}_1 \\ \vdots \\ \bar{v}_n \end{bmatrix}_{n \times m}$$
    $n$ is the number of words, $m$ is the embedding dimension, and $\bar{v}_w$ is the embedding vector for word $w$.
3. Aggregated Word Vectors – The embeddings of words are aggregated to create a text level embedding representation.
    $$v_d = \frac{\sum weight_w \circ \bar{v}_w}{\sum weight_w}$$

$weight_w$ represents the weights for word $w$, and $\bar{v}_w$ represents the embedding vector for word $w$.

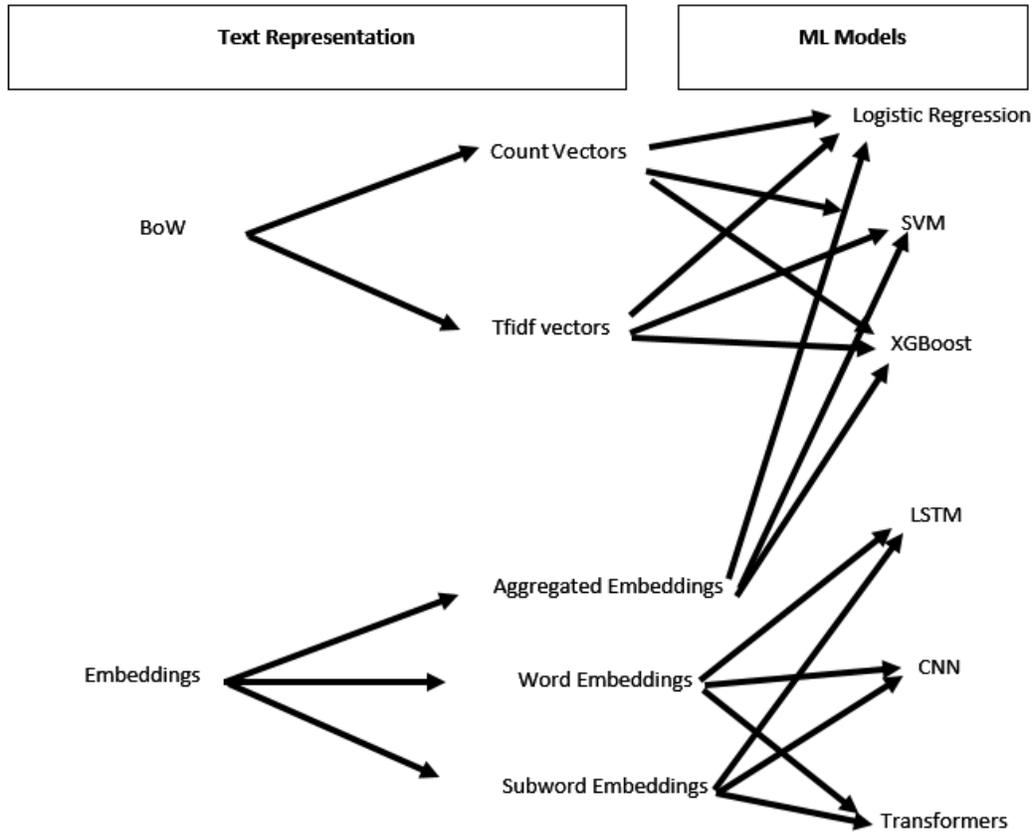

Figure 2-1: Categories of machine learning models (In this paper, we have tested two representations, tf-idf vectors and word embeddings, and three machine learning models SVM, XGBoost and CNN).

## 2.1 Search and Semantically Replace Approach

The method can be broadly divided into two categories, Search, and Semantically Replace. Semantically Replace can be divided further into Replace and Semantic Constraint. The list of different Search, Replace and Semantic Constraint methods for different types of text representations and different NLP models are shown in Figure 2-2.

- **Search**: This is the first step in developing an adversarial text. In this part we find important tokens in the text that can be used to change the model decision. Depending on the type of text representation, we introduce multiple techniques to find important tokens.
- **Replace**: In the next step we provide alternative tokens that can replace the important tokens found in the first step.
- **Semantic Constraint**: The final step can be used to limit the replacements to only those that maintain the semantic and syntactic nature of the text.

We provide more details about each step in the following sections.

### 2.1.1 Search
The following search methods are used to find the "important" part of text:

- Local interpretable model-agnostic explanations (LIME) [16] – We calculate local fidelity - i.e., we extract the explanation that reflects the behavior of the classifier for a particular instance in the form of "important" n-grams. This technique is model-agnostic and is able to explain any model without needing to 'peak' into it.
- Gradient – In this case we back propagate the gradient to the embedding layer and extract the "important" tokens that have the maximum contribution to the gradient of loss w.r.t. the input layer. It indirectly finds the words that maximize the cross-entropy loss.
- Weight based method – In this technique we find the distance of all words from the hyperplane (or the probability logits) from the model.

$$Score(w) = f(h(w))$$

w represents the word, $h(w)$ represents the text representation of the word (i.e., embedding vector, one hot encoding, etc.) and $f$ is the target model.

- Layer-wise relevance propagation (LRP) [17] was recently introduced to explain and interpret DNNs. According to the layer-wise conservation principle, the total relevance at each layer of the neural network are conserved, and in this case we are only concerned with the relevance at the embedding layer.

$$\sum_{i,t} R_{i,t} = \sum_{j,t} R_{j,t} = \sum_{j} R_j = \sum_{k} R_k$$

where $R_k$ is the final prediction of the model before softmax, $R_j$ is the relevance at the maxpool layer, $R_{j,t}$ is the relevance at the convolution layer, and $R_{i,t}$ is the relevance at the embedding layer.

- Tf-idf: Tf-idf values are calculated locally to represent the "importance" of a word in a local text.
- Random: As the name suggests we randomly pick words and replace them.

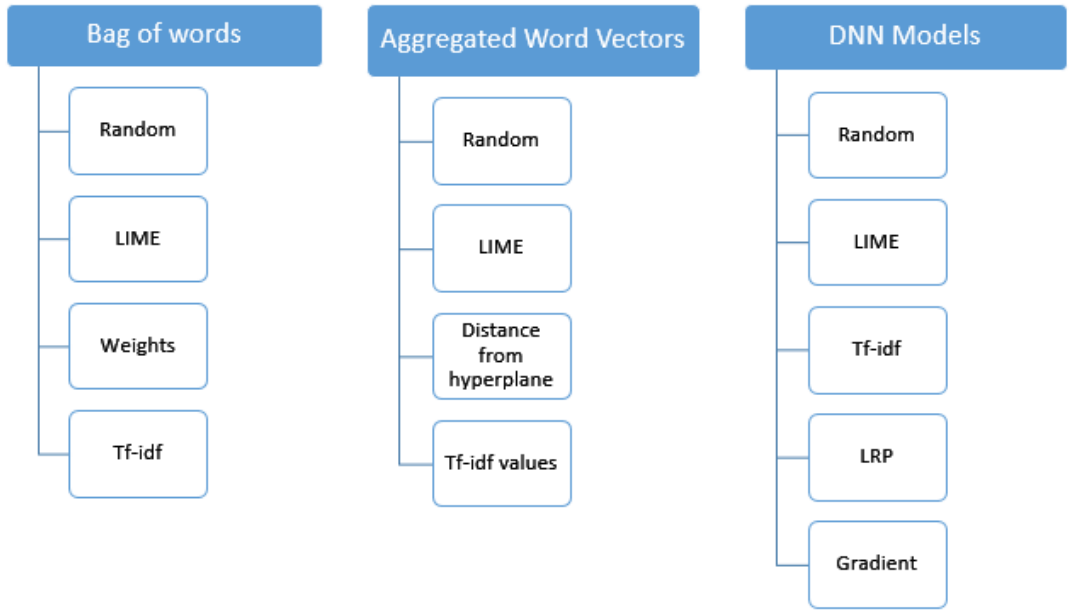

Figure 2-2: Search methods for different representation of text

### 2.1.2 Replace

Following are the replacement methods that are used to find replacements:

- Embedding space – This is used to find the closest words to the "important" words in the embedding space.

$$Score(w') = \cos ine\ similarity(w'_{emb}, w_{emb})$$

  $w'_{emb}$ and $w_{emb}$ represent the embedding vectors of $w'$ and $w$ respectively.
- GPT2 [18] – In this method, we use GPT2 to predict the replacement for word $w$ by feeding $t$ tokens before the word and using GPT2 to complete the sentence.
- Random – This method is to create noise by changing the actual word with minor character modifications. Character perturbations are used to create random changes to a particular word.

In this paper we only apply embedding space strategy to find closest words, and we do not use GPT2 to replace the next word.

### 2.1.3 Semantic Constraint

This part is used to maintain the semantic and syntactic structure of the input text by applying certain constraints on the possible options.

- BERT [15] – BERT's masked language model is used to find the most probable replacement from the set of possible words.
- POS-tag (Part of speech) – Tokens that have the same POS-tag as the original token are used as final attacks.
- Semantic polarity – Semantic polarity of the replacement words should be similar as that of the original word/token.

- Embedding constraint – Only tokens with a minimum cosine similarity are used in the list of final attacks.

The experiments have only been tested using BERT, POS-tag and embedding constraints. The semantic constraint is applied to all experiments that include constraints.

# 3 Data

In this paper, we have used the Consumer Complaint database [19] by Consumer Financial Protection Bureau (CFPB). The Consumer Complaint database is a collection of complaints about consumer financial products and services. CFPB does not verify the allegations in the complaint text and the dataset is updated daily. According to CFPB, complaints are included in the database after the company responds or after 15 days, whichever comes first. In addition, complaints referred to other regulators are not published in the Consumer Complaint database. We downloaded the data from the CFPB website on December 14, 2020 and used a small portion of the data for the research work performed in this paper. Initial CFPB data downloaded in December 2020 contained 1,879,709 complaints. Table 3.1 illustrates an example from the CFPB Consumer Complaint database.

Table 3.1 Example data from the CFPB Consumer Complaint database

| Column Name | Example |
| --- | --- |
| Date received | XX/XX/XXXX |
| Product | Credit reporting, credit repair services, or o… |
| Sub-product | Credit reporting |
| Issue | Incorrect information on your report |
| Sub-issue | Information belongs to someone else |
| Consumer complaint narrative | Consumer XYZ … |
| Company public response | Company has responded to the consumer and the … |
| Company | Company XYZ |
| State | State |
| ZIP code | 123XX |
| Tags | NaN |
| Consumer consent provided? | Consent not provided |
| Submitted via | Web |
| Date sent to company | XX/XX/XXXX |
| Company response to consumer | Closed with explanation |
| Timely response? | Yes |
| Consumer disputed? | No |
| Complaint ID | 123XXXX |

## 3.1 Data Processing

### 3.1.1 Dependent Variable

The product category of the complaint was used as the dependent variable. In this research we built several classification models to categorize the product category of the complaint messages. The original

CFPB database contained 18 product categories as described in Table 3.2. These product categories were aggregated into 4 product categories to simplify the categorization model as shown in column "Final Product Category" of Table 3.2. The final four product categories were used as the dependent variable for the classification model in this research. In addition to combining the product categories, we have also removed any data with missing or NA values for the product categories.

Table 3.2 Product categories of the Consumer Complaint data in the CFPB database

| Initial Product Category | Number of Complaints | Final Product Category |
|---|---|---|
| Prepaid card | 3819 | Card |
| Credit card | 89190 | Card |
| Credit card or prepaid card | 96639 | Card |
| Credit reporting | 140432 | Credit Reporting or Collection |
| Credit repair services, or other personal consumer reports | 582191 | Credit Reporting or Collection |
| Debt collection | 325550 | Credit Reporting or Collection |
| Virtual currency | 18 | Account or Service |
| Bank account or service | 86206 | Account or Service |
| Checking or savings account | 77862 | Account or Service |
| Money transfer, virtual currency, or money service | 21321 | Account or Service |
| Money transfers | 5354 | Account or Service |
| Other financial service | 1059 | Account or Service |
| Payday loan | 5543 | Loan |
| Payday loan, title loan, or personal loan | 15580 | Loan |
| Mortgage | 315948 | Loan |
| Student loan | 59978 | Loan |
| Vehicle loan or lease | 21415 | Loan |
| Consumer Loan | 31604 | Loan |

### 3.1.2 Independent Variable

The *Consumer Complaint Narrative* column, which contains the detailed text of the complaint, was used as the independent variable. The goal of the research was to study the robustness; therefore, the complaint text data was minimally processed. We did not remove any stop words and did not correct for spelling/grammatical mistakes. We only removed data with missing or NA values for the complaint text.

## 3.2 Data Sampling

Initial CFPB data contained 1,879,709 complaints. In order to decrease the computation time, we have decided to use a fraction of initial CFPB data for the modelling purpose. We have selected 17,000 complaints for the train dataset and 2,147 complaints for the test dataset. The counts of complaint texts for the aggregated categories are reported in Table 3.3. It is evident that the dataset is imbalanced, however there are sufficient complaints in each category for text categorization models.

Table 3.3 Final Product Categories of the Consumer Complaint Data

| Final Product Category | Number of Complaints (train) | Number of Complaints (test) |
|---|---|---|
| Account or Service | 1488 | 172 |
| Card | 1810 | 220 |
| Credit Reporting or Collection | 10298 | 1307 |
| Loan | 3404 | 448 |

# 4 Text Classification Modelling Approach

We have considered three machine learning approaches to classify the product categories of the complaints; SVM, XGBoost and CNN. The tf-idf method is used to convert the complaint narrative texts into vectors as inputs to the SVM and XGBoost models. The global vectors for word representation (GloVe) [20] embedding was utilized for training the CNN model. Please see the appendix for additional details. Table 4.1 shows the overall performance of the models on the test dataset. In this paper, we have compared accuracy of the model across various text perturbation schemes.

Table 4.1 Overall performance of the models on the test dataset

| Model | Accuracy |
|---|---|
| **XGBoost** | 0.88 |
| **SVM** | 0.88 |
| **CNN** | 0.87 |

# 5 Result and Discussion

We empirically compute the effectiveness of different combinations of attacks and calculate accuracy as a measure of the potency of the attacks. We start by finding the most potent search method for different machine learning models. Later we combine different constraint methods to create meaningful sentences and compare the different constraints against their effectiveness in reducing the performance of model.

## 5.1 Search

The results of the overall performance drop on the perturbed test set using various search options are shown in Table 5.1. We used GloVe embedding replace option and no constraint (see descriptions in Section 2.1) for the results in Table 5.1. There are notable drops in accuracy for the XGBoost model using LIME or weight based method. However, XGBoost model does not show significant decrease in accuracy using tf-idf or random search approach. This suggests that weight based and LIME based search methods are more effective at findings part of text that contains key unigrams for the BoW XGBoost model. We also changed the fraction of unigrams that were searched and perturbed. With small unigram fraction (<0.1), model accuracy drops by increasing the number of unigrams that is perturbed. However there is no significant decrease in the accuracy once sufficient unigrams (unigram fraction=0.1) have been perturbed. This result suggests that only a small fraction of unigrams can be perturbed to achieve significant model performance deterioration. Therefore computation time can be reduced by capping the number of unigrams that should be perturbed in order to evaluate impact of adversarial testing on the model performance.

Table 5.1 Impact of search methods on XGBoost model performance

| Unigram Fraction | Accuracy (weight) | Accuracy (LIME) | Accuracy (tfidf) | Accuracy (random) |
|---|---|---|---|---|
| baseline | 0.88 | | | |
| 0.05 | 0.79 | 0.80 | 0.87 | 0.87 |
| 0.1 | 0.75 | 0.79 | 0.86 | 0.86 |
| 0.2 | 0.73 | 0.78 | 0.86 | 0.86 |
| 0.3 | 0.73 | 0.76 | 0.86 | 0.86 |

Table 5.2 shows the accuracy drop of the SVM model under various perturbation settings, same as applied for XGBoost. Similar to the XGBoost model, model-based searches of weight and LIME generate more efficient adversarial attacks against SVM, compared to the tf-idf and random perturbations. Between the two model-based searches, the weight search is more efficient than the LIME search; it is because that the weight search is derived from the modeling parameters directly, while the LIME search is conducted through surrogate models thus finding the tokens to be attacked indirectly. In particular, by allowing 20% of tokens to be perturbed, the weight search can generate adversarial texts that reduce accuracy from 88% to 65%, indicating that it is very effective for adversarial attacks.

Table 5.2 Impact of search methods on SVM model performance

| Unigram Fraction | Accuracy (weight) | Accuracy (LIME) | Accuracy (tfidf) | Accuracy (random) |
|---|---|---|---|---|
| baseline | 0.88 | | | |
| 0.05 | 0.80 | 0.84 | 0.87 | 0.87 |
| 0.1 | 0.74 | 0.84 | 0.87 | 0.87 |
| 0.2 | 0.65 | 0.82 | 0.87 | 0.86 |
| 0.3 | 0.62 | 0.82 | 0.86 | 0.86 |

Table 5.3 demonstrates the accuracy drop of the CNN model, with the same unigram perturbation fractions as in Table 5.1. Overall, the CNN model is more vulnerable against adversarial attacks, especially for the three searching strategies, including gradient, LRP and LIME. Similar to the conclusion drawn for the XGBoost model, search methods, such as gradient, LRP and lime, are more efficient than tf-idf based and random perturbation. Among the three attacking approaches, gradient method can result in more significant performance drop. An interesting observation is that for LIME, as the perturbation ratio is increased from 0.1 to 0.3, the accuracy first drops from 0.77 to 0.71, then increases to 0.75. The interpretation is similar as that for the XGBoost and SVM models. The LIME method is based on local surrogate models, therefore may not be accurate in ranking token importance for the CNN model, especially for the words not on the top (i.e., after the unigram fraction increases to a certain point). Similar observation was not drawn from the gradient method. This is not a surprise as the gradient method is approached in the way that it directly identifies the tokens that contribute most to the loss function changes. We can then conclude that gradient method is more efficient in terms of adversarial attacking.

Table 5.3: Impact of search methods on CNN model performance

| Unigram Fraction | Accuracy (gradient) | Accuracy (LRP) | Accuracy (LIME) | Accuracy (tfidf) | Accuracy (random) |
|---|---|---|---|---|---|
| baseline | 0.87 | | | | |
| 0.05 | 0.77 | 0.8 | 0.79 | 0.85 | 0.86 |
| 0.1 | 0.73 | 0.73 | 0.77 | 0.85 | 0.86 |
| 0.2 | 0.72 | 0.71 | 0.71 | 0.83 | 0.83 |
| 0.3 | 0.68 | 0.72 | 0.75 | 0.82 | 0.82 |

## 5.2 Replace

We have focused on embedding based replace option and used three different embeddings to compare performance, "Glove", "w2v" and "w2v-CFPB". The results of the overall performance drop on the perturbed testing set using various replace options are shown in Table 5.4. Per results from Section 5.1, fraction = 0.3 and search method = "weight" yield the most efficient adversarial attack for XGBoost model and SVM model among all conducted tests, and fraction = 0.3 and search method = "gradient" yield the most efficient adversarial attack for the CNN model. We utilized these options and performed a comparison among different replace options, as demonstrated in Table 5.4. No constraint was applied in this set of experiments. Replace option for unigrams using GloVe and w2v embeddings resulted in a larger decrease in model performance compared to w2v-CFPB embeddings and this observation is consistent across all the models. Both GloVe and w2v are developed from the general corpus as provided by the authors of the papers, while w2v-CFPB is internally trained with CFPB's complaint corpus using w2v methodology. The results also suggest that GloVe embeddings generate more efficient adversarial attacks than the w2v embeddings for our testing data and models. Within the same embedding scheme of w2v, the embedding trained from the general corpus is more efficient than the one trained from CFPB's specific corpus. This is because the CFPB specific embedding tends to create replacement unigrams that are from the same domain as the testing data, and hence result in a smaller perturbation in the text vectors compared to the case when the replacement unigrams are chosen from a general domain.

Table 5.4 Impact of replace methods on the model performance

| Replace Embedding | Accuracy (XGBoost) | Accuracy (SVM) | Accuracy (CNN) |
|---|---|---|---|
| baseline | 0.88 | 0.88 | 0.87 |
| GloVe | 0.73 | 0.62 | 0.68 |
| w2v | 0.72 | 0.67 | 0.68 |
| w2v- CFPB | 0.76 | 0.72 | 0.77 |

We performed the same tests of replacement methods for the CNN model, where the gradient searching method and perturbation unigram fraction ratio of 0.3 have been chosen. The result is shown in Table 5.4. The same conclusion was drawn that GloVe embedding as a replacement strategy is more powerful than the w2v-CFPB embedding in terms of attacking efficiency. Differently from the SVM model, the GloVe embedding did not generate as large of a performance drop as the w2v embedding for the CNN model. This observation can be explained by the fact that the embedding layer of the testing CNN model was also initialized with a pre-trained GloVe embedding. When a replacement token is chosen based on the GloVe embedding, it tends to be close to the original token in the same embedding space, thus the input embedding layer of the CNN model has a small perturbation.

## 5.3 Semantic Constraint

The constraint option is used to maintain the semantic and syntactic structure of the text. The results of the overall performance drop on the perturbed testing set using various constraints options are shown in Table 5.5. As per results from Section 5.1, fraction = 0.3 and search method = "weight" yield the most efficient adversarial attack for XGBoost model and SVM model among all conducted tests, and fraction = 0.3 and search method = "gradient" yield the most efficient adversarial attack for the CNN model. As per Section 5.2, GloVe embedding as a replace option is most efficient for all three models. These options were selected and fixed for the comparison in Table 5.5. There are notable drops in accuracy for the XGBoost model using BERT based constraint method, where BERT masked language model is used to find the most potent attack from the replacement set. POS-tag based constraint option uses tokens that have the same POS-tag as the original token for final attacks. Embedding constraint uses tokens with minimum cosine similarity for final attacks. The results in Table 5.5 demonstrate that BERT based constraint option achieves the most performance drop for the XGBoost model.

Table 5.5 Impact of Constraint methods on the model performance

| Constraint | Accuracy (XGBoost) | Accuracy (SVM) | Accuracy (CNN) |
|---|---|---|---|
| baseline | 0.88 | 0.88 | 0.87 |
| No constraint | 0.73 | 0.62 | 0.68 |
| BERT | 0.68 | 0.53 | 0.61 |
| embedding | 0.73 | 0.62 | 0.68 |
| POS-tag | 0.74 | 0.61 | 0.64 |

The comparison of various constraint options when applied on the SVM model is as shown in Table 5.5. Similarly as observed from the XGBoost model setting, BERT constraint achieves the most significant adversarial attacks. Meanwhile, compared to its XGBoost counterpart, the trained SVM model is more sensitive to such adversarial attacks with constraint options.

The same tests using different constraint strategies were conducted on the CNN model, as demonstrated in Table 5.5. Similar to the other models, the largest accuracy drop is observed when the BERT constraint was applied, while the embedding constraint does not cause performance change compared to the case when no constraint was applied. Differently from the XGBoost model and the SVM model, POS-tag constraint tends to cause slight model performance drop compared to the case where no constraint is applied.

# 6  Conclusion

White-box attacks that use the information from the model are more potent in creating adversaries. In the case of XGBoost and SVM it is the weight-based attack, and in the case of CNN it is the gradient-based attacks. There are also a few parts in the text that can cause a significant drop in the model performance; hence, the model prediction is only dependent upon a few words in the text. We show that finding and replacing these sensitive words causes a significant model performance drop, but after the perturbation fraction reaches a certain threshold, adding more perturbations in text does not affect the performance of the model any more. We also show that replacing words using embeddings trained on a different corpus causes more of a performance drop compared to embeddings trained on the same or similar corpus. Lastly, all of the machine learning models are vulnerable to semantic preserving attacks, including CNNs that use embedding space for training.

# Acknowledgements

We would like to acknowledge Harsh Singhal and Vijayan N. Nair for their continuous support and insightful suggestions for the improvement of the manuscript. Furthermore, we are thankful for the help of Agus Sudjianto whose guidance made this research possible.

# A  Appendix A

We considered three machine learning approaches to classify the product categories of the complaints: XGBoost, SVM, and CNN. These three models are described in this section.

## A.1  XGBoost Model with tf-idf

We have created XGBoost model using the tf-idf method to convert the complaint narrative texts into vectors, which are used as input to the model. Since the focus of the paper is on generating adversarial text for machine learning models, we have used pre-optimized hyperparameters for the final XGBoost model, which are shown in Table A.1.1. The model performance was estimated on the testing dataset. The result in Table A.1.2 shows the performance metrics for the predictions on the out of sample testing dataset using the model tuned on the training sample for each category. Table A.1.3 shows the overall performance of the XGBoost model on test dataset. The overall model performance is satisfactory with 0.88 accuracy. In this paper, we have compared accuracy of the model across various text perturbation schemes.

Table A.1.1: Hyper-parameters for XGBoost model

| Hyper-parameter | Value |
|---|---|
| max_depth | 6 |
| min_child_weight | 1 |
| colsample_bytree | 1 |
| Subsample | 1 |
| learning_rate | 0.3 |

Table A.1.2 XGBoost model performance by category on test dataset

| Category | precision | recall | f1-score | support |
|---:|---:|---:|---:|---:|
| Credit Reporting or Collection | 0.91 | 0.93 | 0.92 | 1307 |
| Card | 0.78 | 0.7 | 0.74 | 220 |
| Loan | 0.85 | 0.83 | 0.84 | 448 |
| Account or Service | 0.85 | 0.84 | 0.84 | 172 |

Table A.1.3 Overall performance of XGBoost model on test dataset

|  | Precision | Recall | F1 |
|---|---|---|---|
| **Macro-Average** | 0.85 | 0.82 | 0.83 |
| **Micro-Average** | 0.88 | 0.88 | 0.88 |

| | |
|---|---|
| **Accuracy** | 0.88 |

## A.2  SVM with tf-idf

We developed a second machine learning model with SVM framework. The tf-idf text representation for SVM is similar to that for the XGBoost approach described in Appendix A.1. Specifically, the linear support vector classification is used with L2 penalty, and the regularization parameter is tuned during the model development. The model performance is assessed on the out of sample testing data. Table A.2.1 shows the prediction performance metrics for each separate category; the model performs better with the largest category. Table A.2.2 shows the overall performance of the SVM model on testing data; particularly, the accuracy is 0.88. This SVM model in general achieves very similar performance as the XGBoost model. The simulation studies will compare the model's prediction accuracy across various text perturbation schemes.

Table A.2.1 SVM model performance by category on test dataset

| Category | precision | recall | f1-score | support |
|---:|---:|---:|---:|---:|
| Credit Reporting or Collection | 0.92 | 0.93 | 0.92 | 1307 |
| Card | 0.76 | 0.69 | 0.72 | 220 |
| Loan | 0.84 | 0.85 | 0.84 | 448 |
| Account or Service | 0.84 | 0.84 | 0.84 | 172 |

Table A.2.2 Overall performance of SVM model on test dataset

| | Precision | Recall | F1 |
|---|---|---|---|
| **Macro-Average** | 0.84 | 0.83 | 0.83 |
| **Micro-Average** | 0.88 | 0.88 | 0.88 |
| **Accuracy** | 0.88 | | |

## A.3  CNN model with GloVe embedding

We have utilized a CNN model proposed in a previous work [21]. To utilize information from important single words as well as contexts, the convolutional operation was conducted on unigrams, bigrams, and trigrams of the tokens. The GloVe embedding was utilized for training the model, and the hyper-parameters chosen for the final CNN model, as well as the definition of the parameters, are shown in Table A.3.1.

Table A.3.1: Hyper-parameters for the CNN model

| Hyper-parameter | Value |
|---|---|
| Maximum sequence_length | 256 |
| Batch size | 32 |
| Number of filters | 20 |
| Dropout_embed[2] | 0.5 |
| Dropout_conv[3] | 0.7 |
| Filter Sizes | [1, 2, 3] |
| Embedding Size | 300 |

The CNN model adopts the same train and out-of-time dataset as described in Section 3.2. The CNN model performance on the test set for each category is presented in Table A.3.2, and the overall model performance is presented in Table A.3.3. The CNN model achieved similar level of performance metrics as the XGBoost model (Table A.1.2 and Table A.1.3), with an overall accuracy of 0.87.

Table A.3.2: CNN model performance by category on test dataset

| Category | precision | recall | f1-score | support |
|---:|---:|---:|---:|---:|
| Credit Reporting or Collection | 0.90 | 0.93 | 0.91 | 1307 |
| Card | 0.73 | 0.73 | 0.73 | 220 |
| Loan | 0.84 | 0.80 | 0.82 | 448 |
| Account or Service | 0.84 | 0.77 | 0.80 | 172 |

Table A.3.3: Overall CNN Performance on test dataset

|  | Precision | Recall | F1 |
|---|---|---|---|
| **Macro-Average** | 0.83 | 0.80 | 0.82 |
| **Accuracy** | 0.87 | | |

---

[2] Dropout_embed: Drop-out ratio of the embedding layer
[3] Dropout_conv: Drop-out ratio of the convolutional layers